\documentclass[preprints,article,accept,pdftex,moreauthors]{Definitions/mdpi} 
\firstpage{1} 
\pubvolume{1}
\issuenum{1}
\articlenumber{0}
\pubyear{2026}
\copyrightyear{2026}
\datereceived{ } 
\daterevised{ } % Comment out if no revised date
\dateaccepted{ } 
\datepublished{ } 
\Title{Integer-State Dynamics of Quantized Spiking Neural Networks for Efficient Hardware Acceleration}

\Author{Lei Zhang $^{1}$*\orcidA{}, 
}

\AuthorNames{Lei Zhang}

\address{%
$^{1}$ \quad Faculty of Engineering and Applied Science, University of Regina\\
}

\corres{Correspondence: lei.zhang@uregina.ca}

\abstract{Spiking neural networks (SNNs) support energy-efficient machine intelligence because event-driven computation and sparse activity map naturally to low-power digital hardware. In practical implementations, however, membrane states, synaptic weights, and thresholds are represented with finite-precision integer arithmetic. Quantization, clipping, and overflow can therefore alter network dynamics, not just approximate a higher-precision model. This paper adopts an integer-state dynamical perspective, modeling a hardware-oriented SNN as a deterministic map on a bounded integer lattice. Under this view, recurrence, periodic orbits, and regime changes become intrinsic properties of the system. We introduce a lightweight update rule with integer-valued states and shift-based leakage, and demonstrate the approach through exploratory simulations with network sizes N = 30–130, connection densities 0.1–0.9, and bit widths 4/8/16 over T = 1000 steps. The results show bounded and recurrent temporal structure with strong quantization sensitivity. The observed regimes depend heavily on representation semantics and scaling choices. These findings suggest that numerical precision acts as a dynamical design variable and highlight integer-state analysis as a useful framework for hardware-aware SNN co-design, motivating future work on attractor analysis, precision-aware training, and FPGA/ASIC validation.}


\begin{document}

%%%%%%%%%%%%%%%%%%%%%%%%%%%%%%%%%%%%%%%%%%
\section{Introduction}

The rapid growth of computation- and data- intensive learning systems has created a strong demand for specialized hardware platforms capable of delivering high performance under strict energy and latency constraints \cite{Sze2017,Han2016,Chen2016}. Spiking neural networks (SNNs) provide a promising direction because discrete spike events enable sparse, event-driven computation with intrinsic temporal structure \cite{Maass1997,Pfeiffer2018}. This paradigm links modern machine intelligence with dynamical neuron models from computational neuroscience, offering both computational efficiency and biophysical interpretability \cite{Gerstner2014,Izhikevich2003,zhang2019building,zhang2023neural,zhang2024spike}.

At the hardware level, spike-based computation has been validated across multiple neuromorphic and digital platforms. Large systems such as TrueNorth and SpiNNaker established key architectural foundations for massive parallelism and asynchronous communication \cite{Merolla2014,Furber2014}, while Intel’s Loihi introduced programmable neuron and synapse models with on-chip learning for practical workloads \cite{Davies2018,Davies2021}. In parallel, FPGA- and ASIC-based SNN accelerators exploit integer arithmetic, sparse spike events, and localized synaptic memory to improve efficiency and throughput, forming a broad hardware design space for SNN acceleration \cite{Bouvier2019,Indiveri2015,Esser2016}.

Despite recent progress, the dynamical effects of finite‑precision numerical representations in digital SNNs remain insufficiently studied. In digital implementations, membrane states, synaptic weights, thresholds, and accumulators are represented using bounded integers or fixed-point numbers \cite{Bouvier2019,Indiveri2015}. Consequently, quantization, clipping, and overflow do not simply approximate a higher-precision model; they can alter spike timing, stability, and long-term network dynamics \cite{Hubara2017,Courbariaux2015,Rastegari2016}. 
Therefore, digital SNN implementations form a distinct computational regime whose qualitative behavior strongly depends on bit width, scaling, and update semantics \cite{Roy2019,Schuman2017}.

In this work we adopt an \emph{integer-state dynamical perspective}, modelling a digital SNN as a deterministic dynamical system evolving on a bounded integer lattice. Under this view, trajectories are inherently bounded and eventually recurrent, and thresholding and saturation introduce representation-dependent dynamical behaviors \cite{Indiveri2015,Bouvier2019}. This perspective reframes precision from a source of approximation error to a \emph{dynamical design variable} that determines the network's attractors and activity patterns \cite{Sze2017,Roy2019}. This view also benefits training methods like surrogate gradients by clarifying the constrained state space within which learning must embed network dynamics under quantization \cite{Neftci2019,Zenke2021}.

This paper makes three main contributions. 
First, we formalize an integer-state SNN update rule based on bounded arithmetic and shift-based leakage, allowing the network to be analyzed as a finite discrete-time dynamical system \cite{Schuman2017}. Second, we analyze how quantization, clipping, and overflow affect recurrence and spike patterns organization, revealing strong sensitivity to bit width and scaling \cite{Bouvier2019,Indiveri2015}. Third, we provide exploratory simulations that illustrate representation-dependent dynamical regimes and discuss implications for hardware-aware SNN co-design, particularly for FPGA and ASIC implementations \cite{Esser2016,Roy2019}. These results extend prior work connecting neuron modelling, SNN encoding, and hardware realization, including analytical neuron models \cite{zhang2019building}, bio-inspired dynamics \cite{zhang2023neural, zhang2024spike}, binary and logarithmic spike encoding methods \cite{zhang2025SNNBinaryEncoding, zhang2025LogEncoding}, and FPGA-based SNN implementations \cite{zhang2023spiking,liu2025optimized}.

The remainder of the paper is organized as follows. 
Section~\ref{sec:integer_formulation} introduces the integer-state formulation and update rule. 
Section~\ref{sec:discrete_dynamics} develops the discrete dynamical interpretation. 
Section~\ref{sec:hardware_implications} discusses hardware implications. 
Section~\ref{sec:experiments} presents experimental results and analysis.

%%%%%%%%%%%%%%%%%%%%%%%%%%%%%%%%%%%%%%%%%%
\section{Integer-State SNN Formulation}
\label{sec:integer_formulation}

\subsection{Discrete-Time Neuron Model}

SNNs are often modelled using discrete-time approximations of continuous-time neuron dynamics, particularly for digital hardware or event-driven simulations. Integrate-and-fire (LIF) models remain attractive because they capture accumulation, leakage, thresholding, and spike generation while remaining computationally lightweight \cite{Maass1997,Gerstner2014,Izhikevich2003,Pfeiffer2018}.

Equations~\eqref{eq:lif_update}--\eqref{eq:reset_rule} define a minimal discrete-time integrate-and-fire neuron model suitable for digital implementation. In a generic discrete-time setting, the membrane potential of neuron $i$ evolves according to Eq.~\eqref{eq:lif_update}:

\begin{linenomath}
\begin{equation}
V_i(t+1) = \alpha V_i(t) + \sum_j w_{ij} S_j(t)
\label{eq:lif_update}
\end{equation}
\end{linenomath}

where $V_i(t)$ denotes the membrane potential at time step $t$, $w_{ij}$ represents the synaptic weight from presynaptic neuron $j$ to neuron $i$, and $S_j(t)\in\{0,1\}$ denotes the spike output of neuron $j$. The parameter $\alpha$ models the leakage or decay of the membrane potential over time. A spike is generated whenever the updated membrane potential crosses a threshold $\theta_i$ according to Eq.~\eqref{eq:spike_rule}:

\begin{linenomath}
\begin{equation}
S_i(t+1)=
\begin{cases}
1, & V_i(t+1)\geq \theta_i, \\
0, & \text{otherwise}.
\end{cases}
\label{eq:spike_rule}
\end{equation}
\end{linenomath}

Following spike emission, the membrane potential may be reset or reduced according to a threshold-subtraction rule in Eq.~\eqref{eq:reset_rule}:

\begin{linenomath}
\begin{equation}
V_i(t+1) \leftarrow V_i(t+1)-\theta_i
\label{eq:reset_rule}
\end{equation}
\end{linenomath}

This formulation is intentionally minimal. The purpose of the present paper is not to propose a new biological neuron model, but rather to provide a hardware-oriented state-space abstraction in which the effects of finite precision, bounded arithmetic, and deterministic update semantics can be discussed more directly.

\subsection{Integer-State Representation}

In many software-oriented SNN studies, neuron states and synaptic parameters are represented using floating-point values. In contrast, digital neuromorphic processors, FPGA-based accelerators, and fixed-point SNN pipelines often rely on limited-precision arithmetic in order to reduce memory footprint, simplify arithmetic units, and improve energy efficiency \cite{Davies2018,Indiveri2015,Bouvier2019,Hubara2017}. Motivated by this implementation reality, we consider an integer-state formulation in which membrane potentials, synaptic weights, and thresholds are represented on a bounded integer lattice in Eq.~\eqref{eq:integer_state} :

\begin{linenomath}
\begin{equation}
V_i(t)\in \mathbb{Z}., \quad w_{ij}\in \mathbb{Z}., \quad \theta_i \in \mathbb{Z}.
\label{eq:integer_state}
\end{equation}
\end{linenomath}

Under this assumption, the network update remains structurally similar to the standard discrete-time integrate-and-fire form in Eq.~\eqref{eq:lif_update}, but the state evolution now takes place entirely within a finite or effectively bounded integer state space determined by the bit width and clipping semantics of the implementation. This shift in viewpoint is important: once the neuron states are represented in bounded integer form, the resulting network is no longer merely a floating-point model with quantization noise, but a discrete dynamical system whose reachable trajectories and long-term behavior are shaped by the representation itself. The integer-state constraint in Eq.~\eqref{eq:integer_state} therefore makes the role of bit width explicit and directly links hardware resource limits to state-space size and dynamical behavior.

Such an integer-state representation offers several conceptual and practical advantages. First, it aligns naturally with the arithmetic primitives available in digital hardware, where additions, comparisons, shifts, and bounded accumulations are cheaper than floating-point operations. Second, it makes the role of bit width explicit, thereby connecting hardware resource constraints with state-space size and dynamical behavior. Third, it provides a more suitable abstraction for discussing recurrence, clipping, boundedness, and finite-state effects in hardware-oriented SNNs \cite{Courbariaux2015,Rastegari2016,Hubara2017,Indiveri2015}.

\subsection{Hardware Interpretation}

The integer-state viewpoint is particularly useful when SNNs are mapped to digital hardware. In practice, many neuromorphic and FPGA-based implementations employ fixed-point or integer-valued state variables together with lightweight update operations. A common example is the replacement of a real-valued leakage coefficient by a shift-based approximation in Eq.~\eqref{eq:leak_shift}:

\begin{linenomath}
\begin{equation}
\alpha = 1 - 2^{-k}
\label{eq:leak_shift}
\end{equation}
\end{linenomath}

which yields the hardware-friendly update rule in Eq.~\eqref{eq:shift_update}:

\begin{linenomath}
\begin{equation}
V_i(t+1) = V_i(t) - \left(V_i(t)\gg k\right) + \sum_j w_{ij} S_j(t)
\label{eq:shift_update}
\end{equation}
\end{linenomath}

where $\gg k$ denotes a right-shift by $k$ bits. This form is attractive because it avoids multiplication and can be implemented using only integer subtraction, addition, comparison, and bit-shift operations. In the remainder of this paper, Eq.~\eqref{eq:shift_update} serves as the canonical integer-state update rule used in our simulations.

From a hardware-design perspective, such an update rule reflects the kinds of constraints that appear naturally in digital neuromorphic systems: bounded state variables, quantized weights, finite accumulator widths, and programmable decay rules. Architectures such as Loihi, SpiNNaker, and a variety of FPGA-based SNN accelerators demonstrate that low-precision spike-based computation is not merely an approximation of a higher-precision model, but a practical implementation regime in its own right \cite{Davies2018,Furber2014,Indiveri2015,Bouvier2019}. This interpretation is also consistent with recent work on hardware-oriented SNN design, binary and logarithmic spike encoding, and FPGA-based SNN realization in the author’s research trajectory \cite{zhang2025SNNBinaryEncoding,zhang2025LogEncoding,zhang2023spiking,liu2025optimized}.

Therefore, in this paper, the integer-state SNN formulation should be understood as a hardware-relevant abstraction: a simplified but meaningful representation that captures the essential interaction among quantization, bounded arithmetic, and spike-based recurrence. This abstraction provides the foundation for the discrete dynamical interpretation developed in the next section.

%%%%%%%%%%%%%%%%%%%%%%%%%%%%%%%%%%%%%%%%%%
\section{Discrete Dynamical Perspective}
\label{sec:discrete_dynamics}

\subsection{Integer-State Network as a Discrete Map}
When membrane states, thresholds, and synaptic interactions are represented using bounded integers, a hardware-oriented SNN can be interpreted as a discrete-time state-transition system. Define the vector of neuron states for a network of $N$ neurons at time step $t$ as

\begin{linenomath}
\begin{equation}
\mathbf{X}(t) = \left[V_1(t),V_2(t),\dots,V_N(t)\right]^\top
\label{eq:state_vector}
\end{equation}
\end{linenomath}

Under the integer-state formulation introduced in Section~\ref{sec:integer_formulation}, the network evolves according to the discrete map denoted as

\begin{linenomath}
\begin{equation}
\mathbf{X}(t+1)=F(\mathbf{X}(t))
\label{eq:state_map}
\end{equation}
\end{linenomath}

Here, $F$ is the deterministic update map for implementing leakage, synaptic input accumulation, thresholding, and any reset or clipping oprations. Equations~\eqref{eq:state_vector} and \eqref{eq:state_map} clarify that $F$ is not merely a numerical approximation of a continuous dynamical system; rather, it defines the actual evolution rule of the finite-precision network \cite{Maass1997,Gerstner2014,Indiveri2015,Davies2018}.

This perspective is useful because it shifts the analysis from floating-point neuron trajectories to state-space structure. Rather than evaluating how closely a digital implementation approximates a continuous SNN, we instead study how the finite-state update rule organizes trajectories, recurrence, and long-term dynamical behaviors. Thus the network can be viewed as a quantized dynamical map whose properties depend explicitly on bit width, state bounds, threshold rules, and clipping behavior.

\subsection{Properties of Integer-State Dynamics}

Several important properties follow immediately once the network is treated as a deterministic map on a bounded integer state space. First, if each neuron state is represented with a finite bit width and the update rules enforce bounded arithmetic, then the total number of reachable network states is finite. Second, because the update map $F$ is deterministic, each state has a unique successor. These two properties together imply that every trajectory must eventually revisit a previously visited state. After a state is revisited, the trajectory necessarily enters a periodic cycle. Therefore, with bounded finite-state and deterministic-update, the network must eventually enter either a fixed point or a finite repeating cycle.

However, this statement requires careful interpretation. It is not intended as a complete attractor classification theorem for all quantized SNNs, nor does it claim that all practical networks exhibit short or easily interpretable cycles. Rather, it provides a structural boundary condition: finite-precision SNNs are constrained to evolve within a finite-state graph, and their long-term behavior must be recurrent in a finite-state sense. The key challenges are to characterize how long the transient can be, understand how the recurrence depends on the representation choices, and determine whether the resulting cycles are functionally useful or merely an implementation artifacts \cite{Schuman2017,Roy2019,Hubara2017,Indiveri2015}.

\subsection{Possible Attractor Types}
From this discrete dynamical perspective, the system may exhibit several qualitatively distinct long-term behaviors.

{\bf Fixed points:} 
A fixed point is any state $\mathbf{X}^\star$ satisfying $\mathbf{X}^\star = F(\mathbf{X}^\star)$. In the context of spiking networks, a fixed point may represent complete silence, a clipped steady pattern, or a stable balance among state variables. 

{\bf Periodic orbits:} 
A periodic orbit of period $T$ satisfies $\mathbf{X}(t+T)=\mathbf{X}(t)$ for some positive integer $T$. In hardware SNNs, such periodic behavior may appear as rhythmic spike patterns, recurrent synchronization, or finite state cycles created by quantization and deterministic updates. 

{\bf Long transient dynamics:}
Although a bounded finite-state system must eventually repeat, the transient before recurrence can still be long and complex, especially in higher-dimensional networks with heterogeneous thresholds, sparse connections, and mixed excitatory/inhibitory interactions. From the perspective of practical observation, such transient trajectories may appear irregular or chaos-like over finite windows, even though they remain fundamentally finite-state and ultimately recurrent. The goal here is not to claim formal chaos, but to emphasize that quantized SNNs can still exhibit complex transient dynamics before settling into a repeated orbit.

This taxonomy is useful because it distinguishes three interpretation levels: fixed‑point collapse, periodic recurrence, and long bounded transients that may still be computationally relevant. In hardware systems, all three regimes matter because they lead to different functional and implementation consequences, such as quiescence, oscillation, or sensitivity to scaling and clipping.

\subsection{Conceptual Implications}

Viewing quantized SNNs as finite-state discrete maps has several implications for both theory and hardware design. First, it clarifies that deterministic low-precision implementations are not simply noisy approximations of continuous models, but dynamical systems in their own right. Second, it indicates that precision, clipping, and update rules should be treated as dynamical parameters, not only as engineering constraints. Third, it provides a vocabulary for analyzing recurrence, boundedness, and temporal structure in a way directly relevant to digital implementation.

This perspective also aligns with the author’s broader line of work on neural dynamics and hardware-aware neural modeling. Previous work by the author has examined analytical spiking neuron models \cite{zhang2019building,zhang2023neural}, neural oscillation analysis \cite{zhang2022nnOscillation,zhang2024spike}, and hardware-oriented SNN implementations \cite{liu2025optimized}. Although those works did not directly address the present integer-state formulation, they reinforce the importance of analyzing neural computation through both dynamical structure and implementation semantics.

In this paper, the discrete dynamical perspective serves a conceptual purpose rather than supplying a complete theoretical framework. It provides a principled way to interpret hardware-oriented SNNs as bounded finite-state systems, and motivates the experimental question in Section~\ref{sec:experiments}: whether simple quantized networks exhibit bounded, recurrent, and quantization-sensitive temporal behaviors under lightweight simulation settings.

%%%%%%%%%%%%%%%%%%%%%%%%%%%%%%%%%%%%%%%%%%
\section{Implications for Hardware Design}
\label{sec:hardware_implications}

\subsection{Integer-State Neurons on Digital Hardware}

The integer-state formulation introduced in Section~\ref{sec:integer_formulation} naturally aligns with digital neuromorphic and hardware-accelerated SNN implementations. If membrane states, thresholds, and synaptic weights are represented using bounded integers or fixed-point values, neuron updates reduce to a small set of primitive hardware operations: integer addition, subtraction, comparison, clipping, and bit-shift. Compared with floating-point implementations, this representation avoids expensive arithmetic units and aligns with FPGA logic, ASIC datapaths, and low-power neuromorphic cores \cite{Davies2018,Indiveri2015,Bouvier2019,Furber2014}.

A representative example is the shift-based leakage rule discussed in Section~\ref{sec:integer_formulation}, where the decay parameter is approximated as $\alpha = 1-2^{-k}$ in Eq.~\eqref{eq:leak_shift}. Then, the neuron update follows the integer-state rule in Eq.~\eqref{eq:shift_update}, which can be implemented efficiently using bounded accumulators, right shifts, and threshold comparisons. This computation style is especially attractive for digital spike-processing pipelines because it clarifies the link between representation, arithmetic cost, and temporal behavior. Instead of treating finite precision as an afterthought, the integer-state formulation incorporates it into the computational definition of the model.

\subsection{Benefits of Integer Representation}

The hardware significance of integer‑state SNNs goes beyond ease of implementation, as emphasized in surveys on neuromorphic hardware and efficient deep-learning accelerators \cite{Schuman2017,Bouvier2019,Sze2017}. First, bounded integer states reduce area and energy cost by replacing floating‑point units with simpler datapath operations, consistent with findings in energy-efficient DNN accelerator designs and low-precision computation studies \cite{Sze2017,Chen2016,Hubara2017}. Second, integer bit width explicitly defines the storage memory cost and controls the size of the reachable state space, affecting the system's tendency to saturate, become quiescent, or enter bounded recurrence, as discussed in analyses of neuromorphic circuit behavior and memory constraints \cite{Indiveri2015,Indiveri2011}. Third, integer-state implementations improve timing predictability, aiding pipelined FPGA designs, event schedulers, and real-time neuromorphic execution -- a feature highlighted in hardware neuromorphic platforms such as TrueNorth and Loihi \cite{Merolla2014,Davies2018}. Finally, bounded arithmetic makes clipping and overflow observable and analyzable, allowing these effects to be treated as design variables rather than hidden numerical artifacts, aligning with studies on quantized neural networks, low-precision arithmetic, and neuromorphic information processing constraints \cite{Hubara2017,Sze2017,Indiveri2015,Roy2019}.

These benefits are crucial for hardware-aware SNN design because quantization changes both resource usage and temporal dynamics. A narrow state range may save memory and logic but increase clipping risk and potential collapse of dynamical regimes; a wider state range reduces clipping but may alter recurrence structure and effective scaling. In other words, precision is not solely an engineering trade-off between cost and accuracy. In finite-state SNNs, precision also shapes the dynamical regime of the implemented system.

\subsection{FPGA and ASIC Implementations}

The neuromorphic literature already shows that spike-based computation can be implemented efficiently using finite-precision digital hardware. Examples include TrueNorth \cite{Merolla2014}, SpiNNaker \cite{Furber2014}, and Loihi \cite{Davies2018,Davies2021}. Despite their different architectural philosophies, these systems rely on integer or fixed-point neuron states, programmable decay dynamics, event-driven communication, and bounded local computation to support scalable spike processing. FPGA-based SNN implementations also use bit-serial arithmetic, parallel neuron pipelines, event routing, and sparsity-aware scheduling \cite{Bouvier2019,Indiveri2015}, all of which naturally support integer-state representations.

This hardware perspective is also consistent with the author's recent work on hardware-oriented spiking design and FPGA-based SNN implementation. In particular, recent studies on binary encoding \cite{zhang2025SNNBinaryEncoding}, logarithmic encoding \cite{zhang2025LogEncoding}, FPGA-based SNN implementation \cite{zhang2023spiking}, and hardware-oriented SNN design \cite{liu2025optimized} indicate that representation, computation, and implementation must be considered jointly rather than separately. Although the paper does not yet provide a full FPGA or ASIC realization of the proposed integer-state framework, the conceptual mapping developed here is intended to support such follow-up hardware implementations.

\subsection{Conceptual Takeaways}

From the perspective of this paper, the key hardware lesson is that quantization should be treated as a dynamical design variable. A hardware-oriented SNN is not merely a continuous spiking model implemented approximately with fewer bits. Once bounded integer arithmetic is enforced, the system becomes a finite-state computational object whose reachable trajectories, recurrence structure, and clipping behavior depend directly on representation rules. This observation links the abstract analysis in Section~\ref{sec:discrete_dynamics} with the exploratory simulations in Section~\ref{sec:experiments}. Both suggest that finite-state effects plays a central role in the behavior of  digital SNNs.

The integer-state perspective therefore supports a co-design view of spike-based hardware. The network model influences hardware architecture through bit width, clipping, thresholding, and decay implementation, while hardware constraints feed back into the effective network dynamics. For this paper, the significance of this point is conceptual but important. If finite precision changes the temporal behavior of the network, hardware-aware SNN design should not be viewed merely as a cost-minimization problem. It should also address how implementation choices shape the discrete dynamics of the network.

%%%%%%%%%%%%%%%%%%%%%%%%%%%%%%%%%%%%%%%%%%
\section{Preliminary Experiments}
\label{sec:experiments}

To illustrate the qualitative phenomena discussed in Sections~\ref{sec:integer_formulation}--\ref{sec:hardware_implications}, we performed lightweight exploratory simulations using a finite-state unsigned SNN model without reset -- neurons do not reset their membrane state after spiking. The goal of this section is not a complete theoretical validation, but to demonstrate that bounded finite-state SNNs can exhibit recurrent and quantization-sensitive temporal behavior under practically relevant parameters. The experiments therefore serve as phenomenon demonstrations supporting the conceptual claims of this work.

The global sweep covered network sizes $N=30,32,\dots,130$ and connection densities ranging from $0.1$ to $0.9$ with a step size of $0.1$. The bit-width settings were evaluated from 1 to 16 bits in increments of 1 bit. The lower range (1–4 bits) represents extreme quantization, 8-bit corresponds to typical neuromorphic precision, and 16-bit provides a high-resolution regime approaching near-continuous dynamics. In all experiments, the simulation length was fixed at $T=1000$. Thresholds were sampled from the interval $[4,8]$, and the shift-based leakage parameter was set to $k=1$, corresponding to a decay factor $\alpha=1-2^{-k}=0.5$. In addition to the full sweep, a focused repeated experiment was conducted at $N=64$ and sparsity $0.5$, as this operating region consistently produced bounded recurrent dynamics in exploratory trials. For each bit width from 1 to 16, the same network topology was simulated across 5 predefined random initial conditions to ensure reproducibility.

Figure~\ref{fig1}--Figure~\ref{fig4} provide a focused dynamical view of a representative 8-bit network instance at $N=64$ and sparsity $=0.5$, including the connectivity matrix, membrane-state trajectories, spike raster, and a delay-embedded state visualization. Figure~\ref{fig5} and Figure~\ref{fig6} summarize the effect of bit width across the global sweep. Table~\ref{tab1}, Table~\ref{tab2}, and Table~\ref{tab3} provide numerical summaries supporting the interpretation of these figures.

\subsection{Focused Dynamical Views}

\begin{figure}[H]
\isPreprints{\centering}{} % Only used for preprints
\includegraphics[width=10.0 cm]{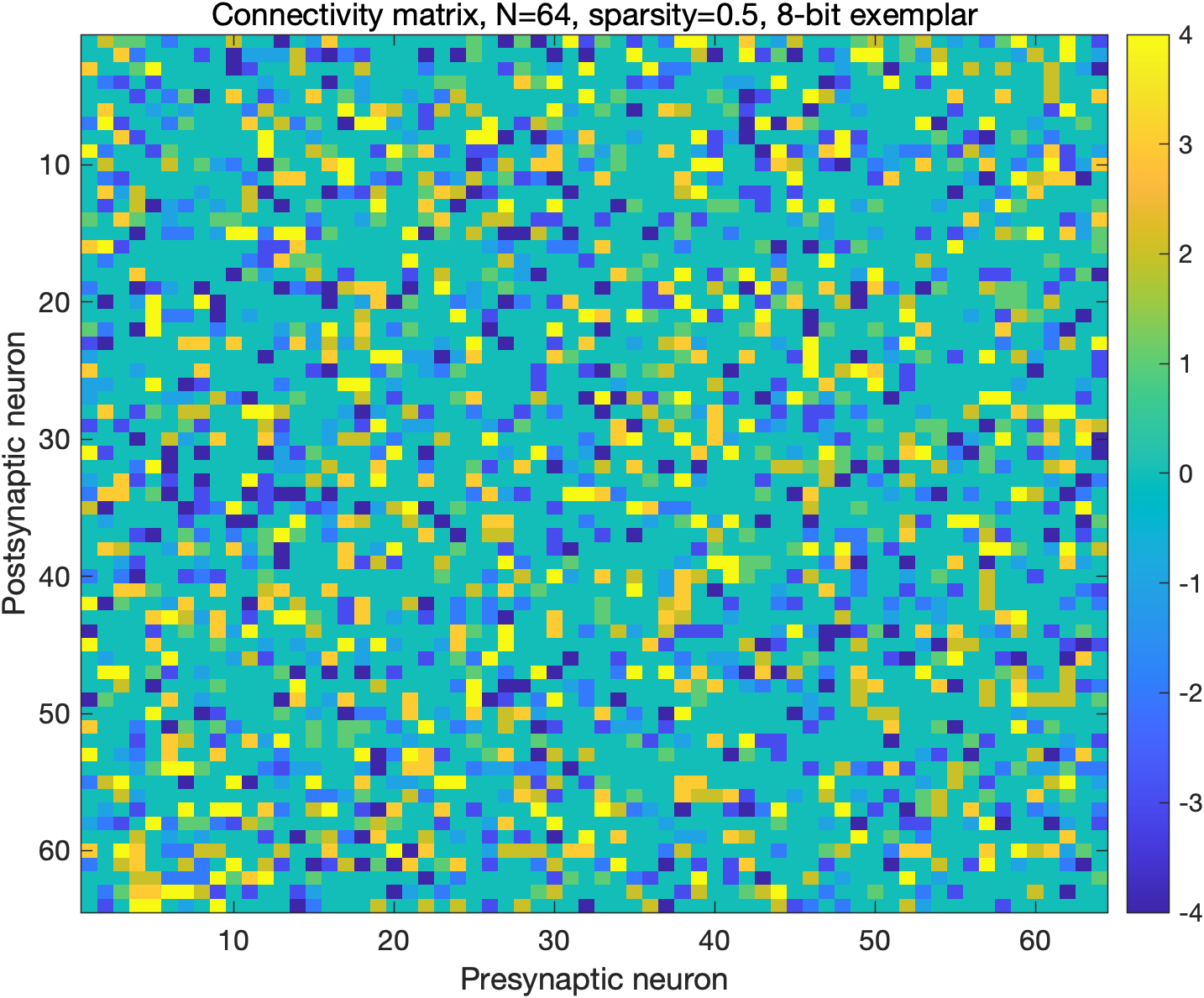}
\caption{Representative sparse integer connectivity matrix for the focused exploratory configuration ($N=64$, sparsity $=0.5$, 8-bit exemplar). The bounded weight structure provides the interaction topology from which the observed finite-state dynamics emerge.\label{fig1}}
\end{figure}

Figure~\ref{fig1} shows the random sparse integer connectivity matrix for a representative 8-bit network instance in the focused configuration. Although the weights are sampled from a simple bounded integer range, the resulting interaction structure can still generate nontrivial temporal behavior under repeated iteration. In this paper, the connectivity matrix is not presented as a learned structure, but as an illustrative finite-state interaction pattern from which the observed dynamics arise.

\unskip

\begin{figure}[H]
\isPreprints{\centering}{} % Only used for preprints
\includegraphics[width=10.0 cm]{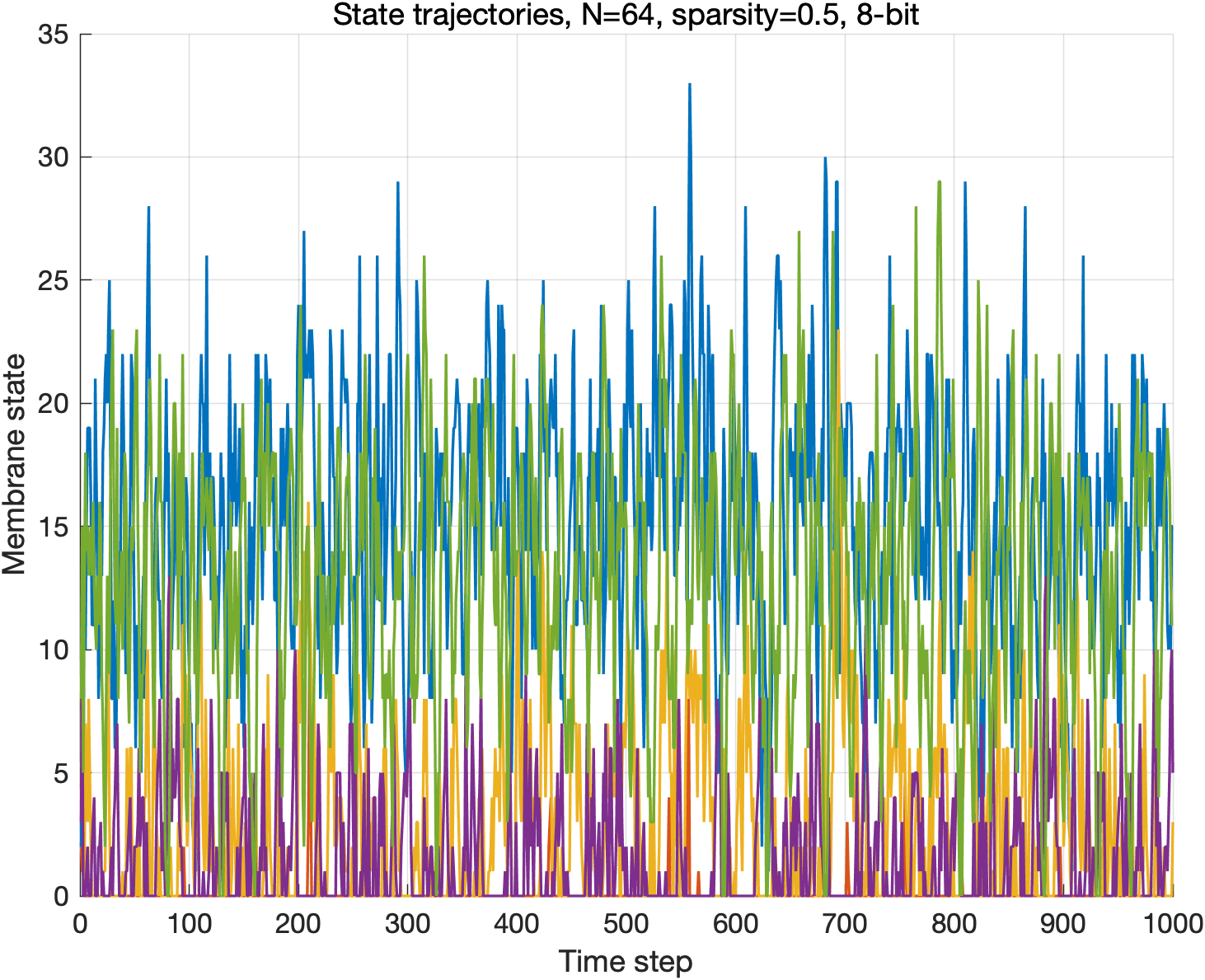}
\caption{Representative membrane-state trajectories for the focused exploratory configuration ($N=64$, sparsity $=0.5$, 8-bit). The trajectories remain bounded and show short-to-moderate recurrent temporal structure.\label{fig2}}
\end{figure}

Figure~\ref{fig2} shows membrane-state trajectories for representative neurons under the same focused configuration. The state variables remain bounded and exhibit repeated but nontrivial temporal structure, rather than diverging or collapsing. This behavior is consistent with the discrete-state viewpoint developed earlier: finite precision constrains the trajectories while still permitting recurrent temporal patterns over finite windows.

\unskip

\begin{figure}[H]
\isPreprints{\centering}{} % Only used for preprints
\includegraphics[width=10.0 cm]{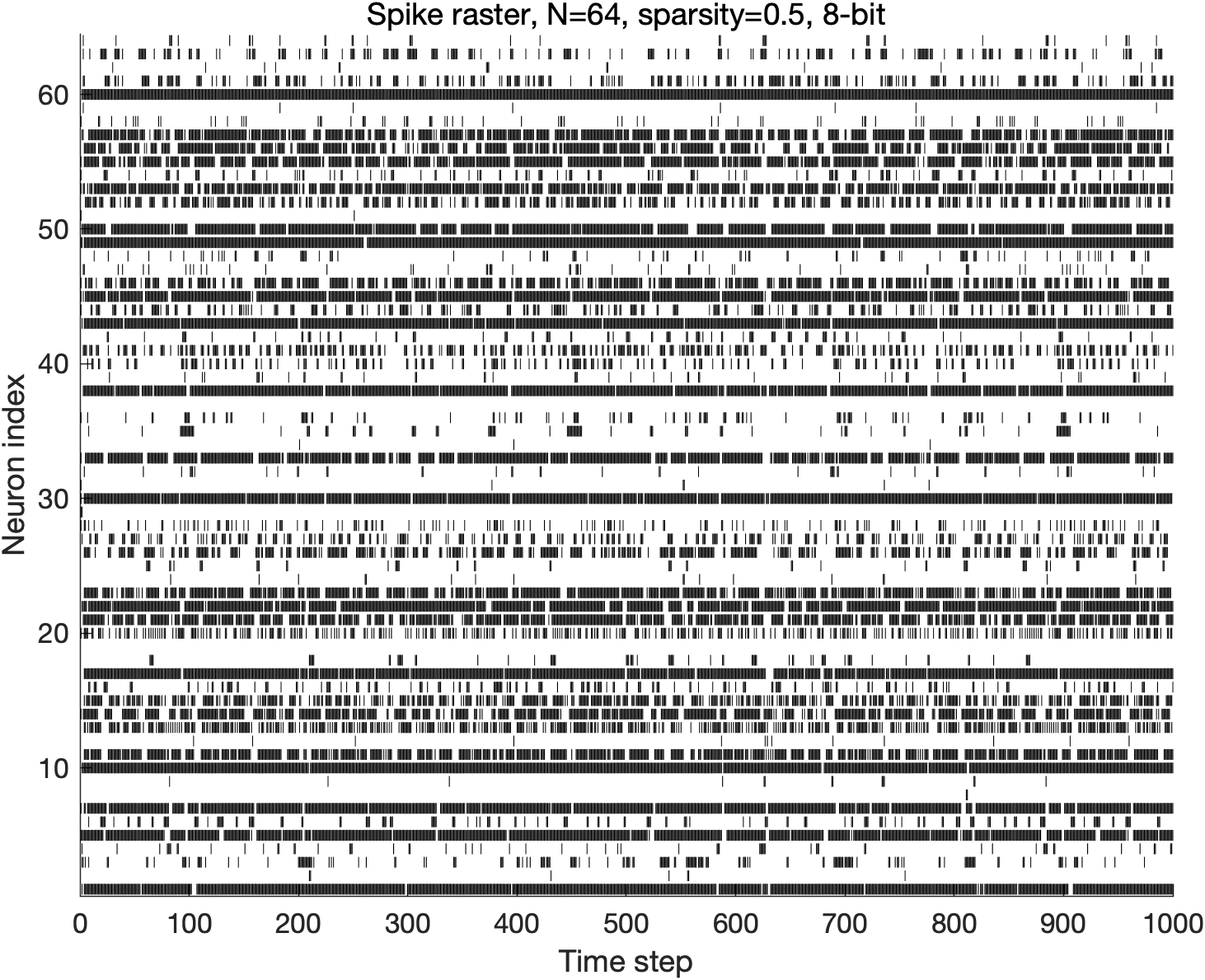}
\caption{Spike raster for the focused exploratory configuration ($N=64$, sparsity $=0.5$, 8-bit). The spike trains exhibit sustained but structured activity rather than immediate silence or saturation.\label{fig3}}
\end{figure}

Figure~\ref{fig3} shows the corresponding spike raster. The network exhibits sustained, structured spike activity throughout the simulation horizon, indicating that the focused regime is neither quiescent nor saturated. Instead, recurrent spike patterns emerge in a bounded state space, representing the qualitative phenomenon targeted by the integer-state interpretation proposed in this work.

\unskip

\begin{figure}[H]
\isPreprints{\centering}{} % Only used for preprints
\includegraphics[width=10.0 cm]{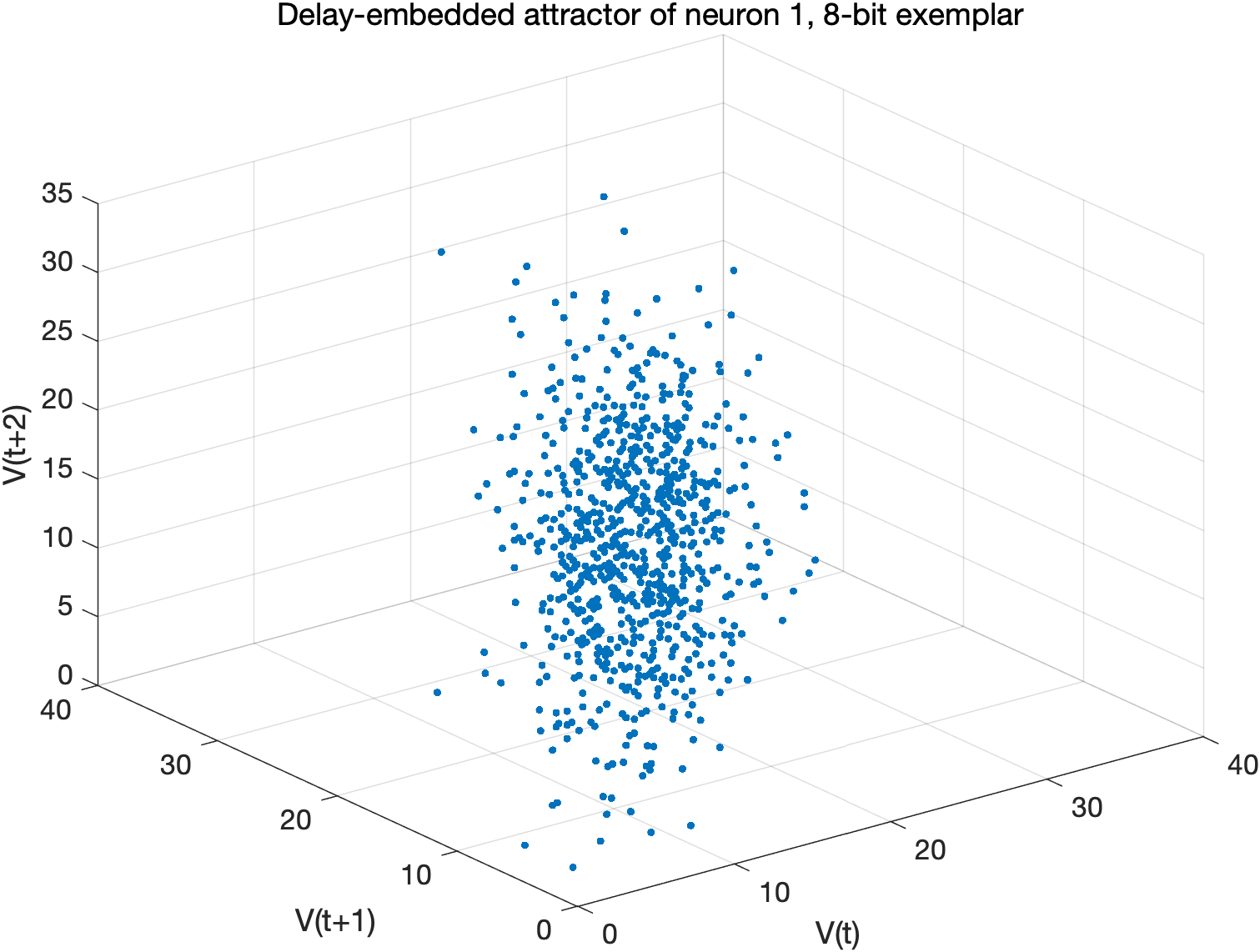}
\caption{Delay-embedded state visualization for a representative neuron in the focused 8-bit configuration. The figure provides an exploratory geometric view of bounded recurrent structure, rather than a formal attractor classification.\label{fig4}}
\end{figure}

Figure~\ref{fig4} shows a delay-embedded visualization of the state evolution for a representative neuron. This figure provides an exploratory geometric view of bounded recurrence rather than a rigorous attractor proof. For this work, this figure serves an interpretive role by making the finite-state temporal structure visually explicit without overstating the theoretical completeness of the analysis.

\unskip

\subsection{Global Bit-Width Effects}

\begin{figure}[H]
\isPreprints{\centering}{} % Only used for preprints
\includegraphics[width=10.0 cm]{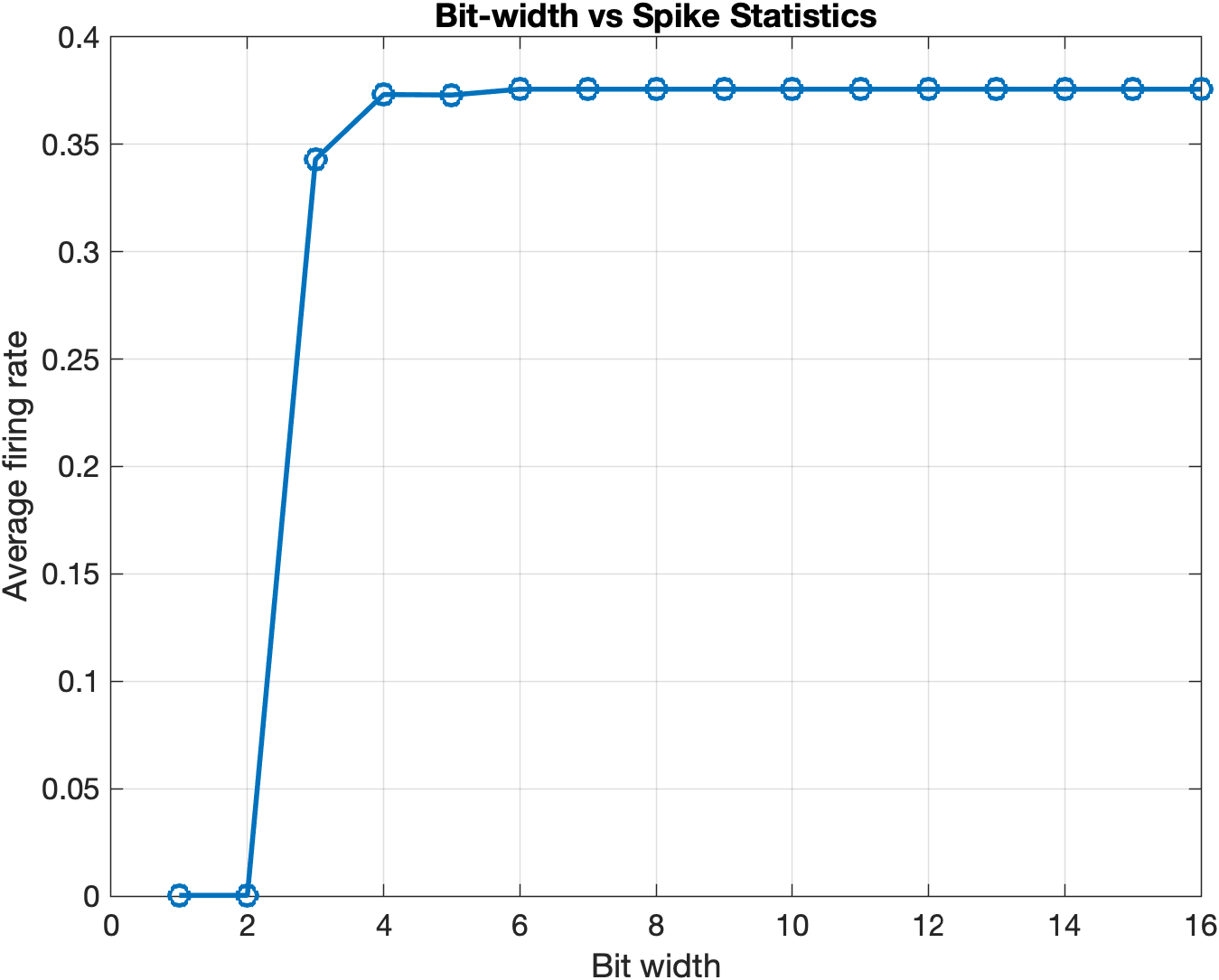}
\caption{Average firing rate as a function of bit width across the global parameter sweep.($k=1$,sparsity$=0.5$)
\label{fig5}}
\end{figure}

\begin{figure}[H]
\isPreprints{\centering}{} % Only used for preprints
\includegraphics[width=10.0 cm]{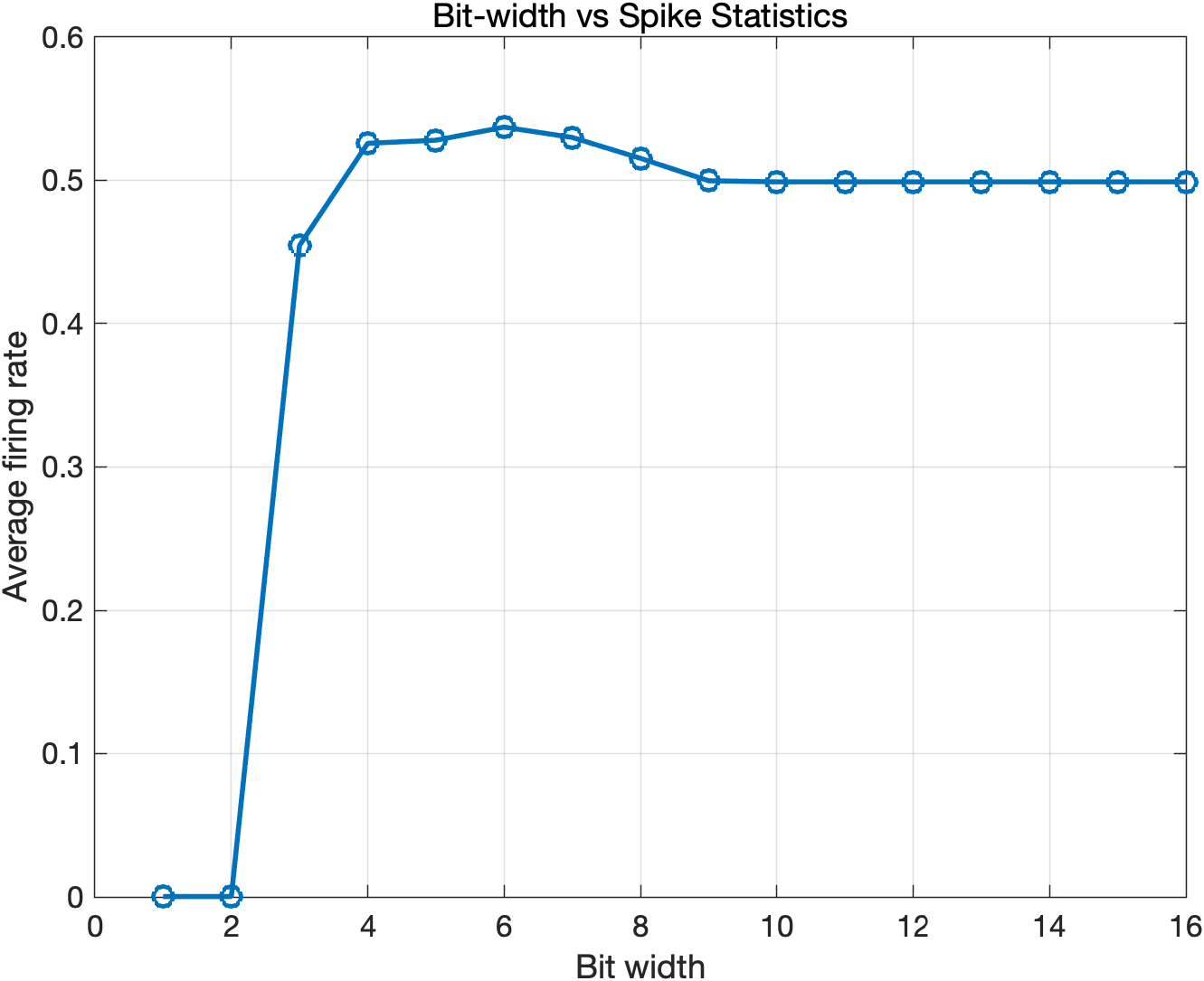}
\caption{Average firing rate as a function of bit width across the global parameter sweep.($k=8$,space$=0.2$)
\label{fig5a}}
\end{figure}

While Figure~\ref{fig1}--Figure~\ref{fig4} focus on a representative network instance, Figure~\ref{fig5} and Figure~\ref{fig5a} summarize the behavior of the global sweep as the bit width changes. 

Figure~\ref{fig5} shows the average firing rate as a function of bit width across the full parameter grid ($k=1$,sparsity$=0.5$). For 1-bit and 2-bit settings, the mean firing rate is effectively zero, indicating a quiescent regime. At 3 bits, activity begins to emerge but remains below the levels observed at higher precision, suggesting a transition near a minimal precision threshold. For bit widths of 4 bits and above, the mean firing rate stabilizes within a narrow range, indicating that sustained activity becomes robust beyond this threshold under the unsigned no-reset formulation. This behavior suggests the existence of a minimal precision threshold separating quiescent and active dynamical regimes in integer-state SNNs.

\unskip

Figure~\ref{fig5a} shows the corresponding relationship for a ddifferent parameter setting ($k=8$,sparsity$=0.2$).While both settings exhibit a similar transition from quiescent to active regimes around 3–4 bits, the structure of the active regime differs. For $k=1$ and sparsity $0.5$, the firing rate quickly converges to a stable plateau, indicating a precision-insensitive regime. In contrast, for $k=8$ and sparsity $0.2$, the plateau exhibits measurable variation across bit widths, suggesting that longer memory and reduced connectivity increase sensitivity to quantization.

These results indicate that precision acts as a dynamical control variable whose effect depends on the interaction between leakage and network sparsity.

\unskip

\begin{figure}[H]
\isPreprints{\centering}{} % Only used for preprints
\includegraphics[width=10.0 cm]{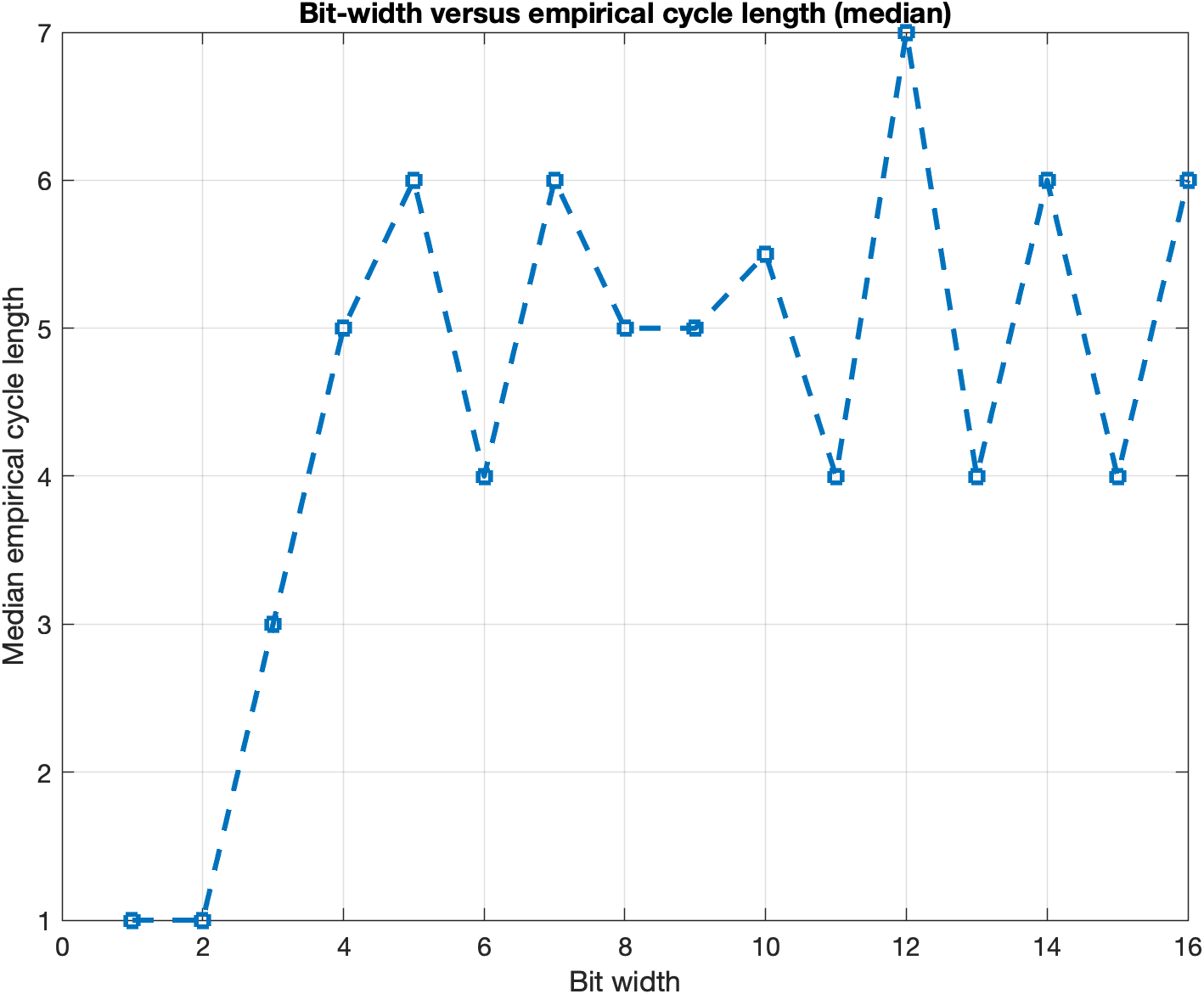}
\caption{Median empirical cycle length as a function of bit width across the global parameter sweep. Bounded recurrence appears across multiple precision regimes rather than being confined to a single bit-width setting.\label{fig6}}
\end{figure}

Figure~\ref{fig6} shows the median empirical cycle length across the global sweep. The results indicate that bounded recurrence is not restricted to one precision setting. Instead, recurrent or short repeating temporal regimes appear across multiple bit-width choices, supporting the view that integer-state SNNs should be studied as discrete dynamical systems in their own right.

\unskip

\subsection{Quantitative Summary}

The numerical summaries reinforce the qualitative patterns shown in Figure~\ref{fig1}--Figure~\ref{fig6}. Table~\ref{tab1} reports the global sweep summary by bit width. The mean firing rate remains close to $0.36$ across 4-, 8-, and 16-bit settings, while the mean pseudo-rank surrogate and the median empirical cycle length remain nontrivial. These results indicate that the exploratory model frequently supports bounded recurrent temporal structure over the explored parameter space.

\begin{table}[H]
\caption{Global sweep summary by bit width. The results indicate that bounded recurrent and quantization-sensitive dynamics are observed across all three tested precision levels.\label{tab1}}
\begin{tabularx}{\textwidth}{CCCCC}
\toprule
\textbf{Bits} & \textbf{Mean Firing Rate} & \textbf{Active Neuron Fraction (Mean)} & \textbf{Pseudo-rank (Mean)} & \textbf{Empirical Cycle (Median)}\\
\midrule
4  & 0.3646 & 0.5883 & 44.1634 & 4.5\\
8  & 0.3629 & 0.5597 & 41.6057 & 5.5\\
16 & 0.3626 & 0.5715 & 43.5708 & 5.0\\
\bottomrule
\end{tabularx}
\end{table}

Table~\ref{tab2} focuses on the repeated experiment at $N=64$ and sparsity $=0.5$. In this region, the mean firing rates remain around $0.40$, and the median empirical recurrence period lies in the range of approximately 8--10.5 time steps. This confirms that the focused operating region repeatedly produces visually rich bounded recurrence under the exploratory update semantics.

\begin{table}[H]
\caption{Focused repeated results for the exploratory configuration $N=64$ and sparsity $=0.5$. This region repeatedly produced visually rich bounded recurrent behavior.\label{tab2}}
\begin{tabularx}{\textwidth}{CCCCC}
\toprule
\textbf{Bits} & \textbf{Mean Firing Rate} & \textbf{Std Firing Rate} & \textbf{Pseudo-rank (Mean)} & \textbf{Empirical Cycle (Median)}\\
\midrule
4  & 0.4068 & 0.0678 & 27.4667 & 10.5\\
8  & 0.4179 & 0.0759 & 25.1333 & 10.0\\
16 & 0.4045 & 0.0672 & 27.9000 & 8.0\\
\bottomrule
\end{tabularx}
\end{table}

Table~\ref{tab3} lists representative top recurrent cases from the global sweep according to the pseudo-rank surrogate. These examples suggest that larger networks with moderate connection density can produce especially rich tail dynamics under the exploratory unsigned integer-state update rule.

\begin{table}[H]
\caption{Representative top recurrent cases from the global sweep, ranked by the pseudo-rank surrogate. These examples suggest that moderately connected larger networks often exhibit richer bounded tail dynamics in the exploratory regime.\label{tab3}}
\begin{tabularx}{\textwidth}{CCCCC}
\toprule
\textbf{N} & \textbf{Sparsity} & \textbf{Bits} & \textbf{Pseudo-rank} & \textbf{Firing Rate}\\
\midrule
130 & 0.2 & 8  & 126 & 0.3588\\
128 & 0.2 & 4  & 126 & 0.3206\\
130 & 0.3 & 4  & 124 & 0.3877\\
130 & 0.2 & 16 & 124 & 0.3635\\
\bottomrule
\end{tabularx}
\end{table}

\subsection{Interpretation}

Taken together, Figure~\ref{fig1}--Figure~\ref{fig6} and Table~\ref{tab1}--Table~\ref{tab3} support three practical conclusions. First, finite-state integer SNNs can exhibit bounded and recurrent temporal behavior over broad regions of parameter space, even without learning or external drive. Second, bit width should not be interpreted only as a storage or arithmetic parameter, since it also interacts with clipping, effective dynamic range, and recurrence structure. Third, the particular region around $N=64$ and sparsity $=0.5$ appears to provide a useful operating point for visually exposing bounded recurrent behavior in the exploratory regime. Although the present experiments are intentionally lightweight and phenomenological, they support the central argument of this paper: finite precision and integer-state constraints are not merely implementation details, but factors that can materially shape the observed temporal regime of hardware-oriented SNNs.

%%%%%%%%%%%%%%%%%%%%%%%%%%%%%%%%%%%%%%%%%%
\section{Discussion}
\label{sec:discussion}

This paper argues that quantized SNNs should be modeled as finite-state dynamical systems whose behavior is governed by discrete state transitions rather than continuous approximations. Under bounded integer representations, the system state space is finite, implying that trajectories must eventually enter periodic orbits or fixed points. Therefore, recurrence is not an empirical artifact but a structural property of the model. The focused six‑figure analysis shows that even lightweight integer‑state SNNs can display structured bounded trajectories under simple update rules, while the global sweep confirms that recurrent activity persists across 4-, 8-, and 16‑bit regimes. Thus, precision selection should be discussed jointly with temporal behavior, not only with hardware cost.

Several limitations follow naturally. The current experiments are designed to expose qualitative dynamical regimes rather than to exhaustively characterize attractor structures. The unsigned/no‑reset regime highlights recurrent structure but differs from the stricter signed/reset formulation, emphasizing how update semantics strongly influence dynamics. Earlier “attractor length” metrics were also revised into clearer empirical indicators—pseudo‑rank and repeated‑tail recurrence—to avoid misinterpretation as formal dynamical invariants. This integer‑state view aligns with the author’s broader work on spiking models, neural dynamics, encoding, and FPGA‑oriented SNN design, and provides a conceptual bridge between dynamics, low‑precision computation, and neuromorphic implementation. A formal characterization of orbit length distributions as a function of bit width remains an open problem.

Future directions include: (1) rigorous analysis of recurrence classes under different reset/clipping semantics; (2) FPGA/ASIC implementations with measured resource and energy behavior; (3) interactions between integer‑state recurrence and training, encoding, or task‑level performance. These steps would move the present work toward a principled hardware–algorithm co‑design framework for low‑precision spike‑based computation.

%%%%%%%%%%%%%%%%%%%%%%%%%%%%%%%%%%%%%%%%%%
\section{Conclusions}
\label{sec:conclusion}

This work establishes that quantization is not merely a precision constraint but a mechanism that fundamentally alters the dynamical class of spiking neural networks. By enforcing a finite state space, integer representations guarantee the emergence of periodic orbits and bounded trajectories, redefining how stability and temporal behavior should be analyzed in SNNs.The next steps include: formal analysis of finite‑state orbits, FPGA/ASIC implementations, and studying how integer‑state recurrence interacts with training, encoding, and performance. This perspective suggests that future SNN design should treat precision, state bounds, and update semantics as primary dynamical parameters, enabling principled co-design of algorithms and hardware.

%%%%%%%%%%%%%%%%%%%%%%%%%%%%%%%%%%%%%%%%%%
\vspace{6pt} 

%\funding{This research was funded by the Natural Sciences and Engineering Research Council of Canada (NSERC), Discovery Development Grant (DDG-2024-00034). Additional support was provided by the University of Regina.}

%%%%%%%%%%%%%%%%%%%%%%%%%%%%%%%%%%%%%%%%%%
\isPreprints{}{% This command is only used for ``preprints''.
\begin{adjustwidth}{-\extralength}{0cm}
} % If the paper is ``preprints'', please uncomment this parenthesis.
%\printendnotes[custom] % Un-comment to print a list of endnotes

\reftitle{References}

% Please provide either the correct journal abbreviation (e.g. according to the “List of Title Word Abbreviations” http://www.issn.org/services/online-services/access-to-the-ltwa/) or the full name of the journal.
% Citations and References in Supplementary files are permitted provided that they also appear in the reference list here. 

%=====================================
% References, variant A: external bibliography
%=====================================
\bibliography{references_intro_aligned260314}

%=====================================
% References, variant B: internal bibliography
%=====================================

% If authors have biography, please use the format below
%\section*{Short Biography of Authors}
%\bio
%{\raisebox{-0.35cm}{\includegraphics[width=3.5cm,height=5.3cm,clip,keepaspectratio]{Definitions/author1.pdf}}}
%{\textbf{Firstname Lastname} Biography of first author}
%
%\bio
%{\raisebox{-0.35cm}{\includegraphics[width=3.5cm,height=5.3cm,clip,keepaspectratio]{Definitions/author2.jpg}}}
%{\textbf{Firstname Lastname} Biography of second author}

% For the MDPI journals use author-date citation, please follow the formatting guidelines on http://www.mdpi.com/authors/references
% To cite two works by the same author: \citeauthor{ref-journal-1a} (\citeyear{ref-journal-1a}, \citeyear{ref-journal-1b}). This produces: Whittaker (1967, 1975)
% To cite two works by the same author with specific pages: \citeauthor{ref-journal-3a} (\citeyear{ref-journal-3a}, p. 328; \citeyear{ref-journal-3b}, p.475). This produces: Wong (1999, p. 328; 2000, p. 475)

%%%%%%%%%%%%%%%%%%%%%%%%%%%%%%%%%%%%%%%%%%
%% for journal Sci
%\reviewreports{\\
%Reviewer 1 comments and authors’ response\\
%Reviewer 2 comments and authors’ response\\
%Reviewer 3 comments and authors’ response
%}
%%%%%%%%%%%%%%%%%%%%%%%%%%%%%%%%%%%%%%%%%%
%\PublishersNote{}
\isPreprints{}{% This command is only used for ``preprints''.
\end{adjustwidth}
} % If the paper is ``preprints'', please uncomment this parenthesis.
\end{document}